\documentclass[runningheads]{llncs}
\usepackage{graphicx}

\usepackage{multirow}
\usepackage{booktabs}
\usepackage{tabularx}
\usepackage{amsmath}
\usepackage{subcaption}
\usepackage{hyperref}
\usepackage{orcidlink}
\usepackage{color,soul}
\usepackage{diagbox}
\usepackage{upgreek}
\newlength{\twosubht}
\newsavebox{\twosubbox}

\begin{document}

\newcommand{\paperTodo}[1]{\textcolor{red}{\textbf{\hl{#1}}}}

\title{Many tasks make light work: Learning to localise medical anomalies from multiple synthetic tasks}
\titlerunning{Many tasks make light work}
%
%

\author{Matthew Baugh\inst{1}\orcidlink{0000-0001-6252-7658} \and
Jeremy Tan\inst{2}\orcidlink{0000-0002-9769-068X} \and
Johanna P. M\"uller\inst{3}\orcidlink{0000-0001-8636-7986} \and
Mischa Dombrowski\inst{3}\orcidlink{0000-0003-1061-8990} \and
James Batten\inst{1}\orcidlink{0000-0002-8028-5709} \and
Bernhard Kainz\inst{1,3}\orcidlink{0000-0002-7813-5023}}

\authorrunning{M. Baugh et al.}

\institute{Imperial College London, UK \\
\email{matthew.baugh17@imperial.ac.uk}
\and ETH Zurich, CH
\and Friedrich--Alexander University Erlangen--N\"urnberg, DE}

\maketitle              
\begin{abstract}
There is a growing interest in single-class modelling and out-of-distribution detection as fully supervised machine learning models cannot reliably identify classes not included in their training. 
The long tail of infinitely many out-of-distribution classes in real-world scenarios, \emph{e.g.}, for screening, triage, and quality control, means that it is often necessary to train single-class models that represent an expected feature distribution, \emph{e.g.}, from only strictly healthy volunteer data. 
Conventional supervised machine learning would require the collection of datasets that contain enough samples of all possible diseases in every imaging modality, which is not realistic. 
Self-supervised learning methods with synthetic anomalies are currently amongst the most promising approaches, alongside generative auto-encoders that analyse the residual reconstruction error.
However, all methods suffer from a lack of structured validation, which makes calibration for deployment difficult and dataset-dependant.
Our method alleviates this by making use of multiple visually-distinct synthetic anomaly learning tasks for both training and validation.
This enables more robust training and generalisation.
With our approach we can readily outperform state-of-the-art methods, which we demonstrate on exemplars in brain MRI and chest X-rays.
Code is available at \href{https://github.com/matt-baugh/many-tasks-make-light-work}{https://github.com/matt-baugh/many-tasks-make-light-work}.
\end{abstract}
\section{Introduction}
\parskip0pt
In recent years, the workload of radiologists has grown drastically, quadrupling from 2006 to 2020 in Western Europe \cite{bruls2020workload}.
This huge increase in pressure has led to long patient-waiting times and fatigued radiologists who make more mistakes~\cite{Brady2017-tf}.
The most common of these errors is underreading and missing anomalies (42\%); followed by missing additional anomalies when concluding their search after an initial finding (22\%) \cite{doi:10.2214/AJR.13.11493}.
Interestingly, despite the challenging work environment, only 9\% of errors reviewed in \cite{doi:10.2214/AJR.13.11493} were due to mistakes in the clinicians' reasoning.
Therefore, there is a need for automated second-reader capabilities, which brings any kind of anomalies to the attention of radiologists.
For such a tool to be useful, its ability to detect rare or unusual cases is particularly important.
Traditional supervised models would not be appropriate, as acquiring sufficient training data to identify such a broad range of pathologies is not feasible.
Unsupervised or self-supervised methods to model an expected feature distribution, \emph{e.g.}, of healthy tissue, is 
therefore a more natural path, as they are geared towards identifying any deviation from the normal distribution of samples, rather than a particular type of pathology. 

There has been rising interest in using end-to-end self-supervised methods for anomaly detection.
Their success is most evident at the MICCAI Medical Out-of-Distribution Analysis (MOOD) Challenge \cite{zimmerer2022mood}, where all winning methods have followed this paradigm so far (2020-2022).
These methods use the variation within normal samples to generate diverse anomalies through sample mixing \cite{melba:2022:013:tan,PII,mood_2021_winner,NSA}.
However all these methods lack a key component: structured validation.
This creates uncertainty around the choice of hyperparameters for training.
For example, selecting the right training duration is crucial to avoid overfitting to proxy tasks.
Yet, in practice, training time is often chosen arbitrarily, reducing reproducibility and potentially sacrificing generalisation to real anomalies.

\noindent\textbf{Contribution: } 
We propose a cross-validation framework, using separate self-supervision tasks to minimise overfitting on the synthetic anomalies that are used for training.
To make this work effectively we introduce a number of non-trivial and seamlessly-integrated synthetic tasks, each with a distinct feature set so that during validation they can be used to approximate generalisation to unseen, real-world anomalies.
To the best of our knowledge, this is the first work to train models to directly identify anomalies on tasks that are deformation-based, tasks that use Poisson blending with patches extracted from external datasets, and tasks that perform efficient Poisson image blending in 3D volumes, which is in itself a new contribution of our work.
We also introduce a synthetic anomaly labelling function which takes into account the natural noise and variation in medical images.
Together our method achieves an average precision score of 76.2 for localising glioma and 78.4 for identifying pathological chest X-rays, thus setting the state-of-the-art in self-supervised anomaly detection.

\noindent\textbf{Related Work: }
The most prevalent methods for self-supervised anomaly detection are based on generative auto-encoders that analyse the residual error from reconstructing a test sample.
This is built on the assumption that a reconstruction model will only be able to correctly reproduce data that is similar to the instances it has been trained on, \emph{e.g.} only healthy samples.
Theoretically, at test time, the residual reconstruction error should be low for healthy tissues but high for anomalous features. 
This is an active area of research with several recent improvements upon the initial idea~\cite{SCHLEGL201930}, \emph{e.g.}, 
\cite{pinaya2022fast} applied a diffusion model to a VQ-VAE \cite{van2017neural} to resample the unlikely latent codes and  
\cite{zhang2022multi} gradually transition from a U-Net architecture to an autoencoder over the training process in order to improve the reconstruction of finer details.
Several other methods aim to ensure that the model will not reproduce anomalous regions by training it to restore samples altered by augmentations such as masking out regions \cite{zimmerer2018context}, interpolating heavily augmented textures \cite{zavrtanik2021draem} or adding coarse noise \cite{kascenas2023role}.
\cite{cai2022dual} sought to identify more meaningful errors in image reconstructions by comparing the reconstructions of models trained on only healthy data against those trained on all available data.

However, the general assumption that reconstruction error is a good basis for an anomaly scoring function has recently been challenged.
Auto-encoders are unable to identify anomalies with extreme textures~\cite{meissen2022pitfalls}, are reliant on empirical post-processing to reduce false-positives in healthy regions~\cite{baur2021autoencoders} and can be outperformed by trivial approaches like thresholding of FLAIR MRI~\cite{meissen2022challenging}.

Self-supervised methods take a more direct approach, training a model to directly predict an anomaly score using synthetic anomalies.
Foreign patch interpolation (FPI) \cite{melba:2022:013:tan} was the first to do this at a pixel-level, by linearly interpolating patches extracted from other samples and predicting the interpolation factor as the anomaly score.
Similar to CutPaste~\cite{li2021cutpaste}, \cite{mood_2021_winner} fully replaces 3D patches with data extracted from elsewhere in the same sample, but then trains the model to segment the patches.
 Poisson image interpolation (PII)~\cite{PII} seamlessly integrates sample patches into training images, preventing the models from learning to identify the anomalies by their discontinuous boundaries.
Natural synthetic anomalies (NSA)~\cite{NSA} relaxes patch extraction to random locations in other samples and introduces an anomaly labelling function based on the changes introduced by the anomaly.

Some approaches combine self-supervised and reconstruction-based methods by training a discriminator to compute more exact segmentations from reconstruction model errors \cite{zavrtanik2021draem,cai2022dual-extension}.
Other approaches have also explored contrasting self-supervised learning for anomaly detection \cite{luth2023cradl,CCDmiccai2021}.

\section{Method}

The core idea of our method is to use synthetic tasks for both training and validation.
This allows us to monitor performance and prevent overfitting, all without the need for real anomalous data.
Each self-supervised task involves introducing a synthetic anomaly into otherwise normal data whilst also producing the corresponding label.
Since the relevant pathologies are unknown a priori, we avoid simulating any specific pathological features. 
Instead, we use a wide range of subtle and well-integrated anomalies to help the model detect many different kinds of deviations, ideally including real unforeseen anomalies.   
In our experiments, we use five tasks, but more could be used as long as each one is sufficiently unique.
Distinct tasks are vital because we want to use these validation tasks to estimate the model's generalisation to unseen classes of anomalies.
If the training and validation tasks are too similar, the performance on the validation set may be an overly optimistic estimate of how the model would perform on unseen real-world anomalies.

When performing cross-validation over all synthetic tasks and data partitions independently, the number of possible train/validation splits increases significantly, requiring us to train $F\cdot(T_{N}\mathbf{C}T)$ independent models, where $T_{N}$ is the total number of tasks, $T$ is the number of tasks used to train each model and $F$ is the number of data folds, which is computationally expensive.
Instead, as in our case $T_{N}{=}F{=}5$, we opt to associate each task with a single fold of the training data (Figure~\ref{fig:cross_val}).
We then apply $5\mathbf{C}T$-fold cross-validation over each combination.
In each iteration, the corresponding data folds are collected and used for training or validation, depending on which partition forms the majority.

\sbox\twosubbox{%
  \resizebox{.95\linewidth}{!}{%
    \includegraphics[height=3cm]{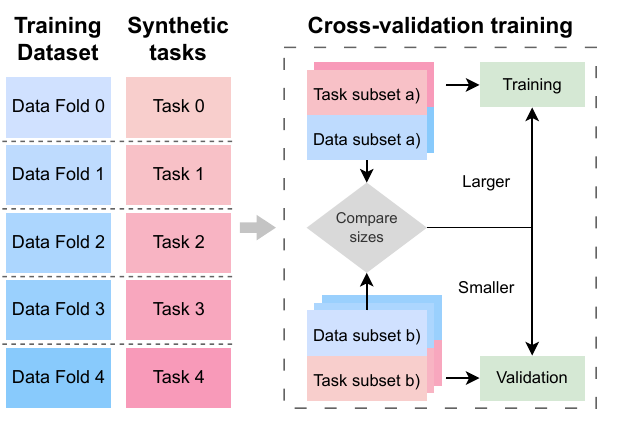}%
    \includegraphics[height=3cm]{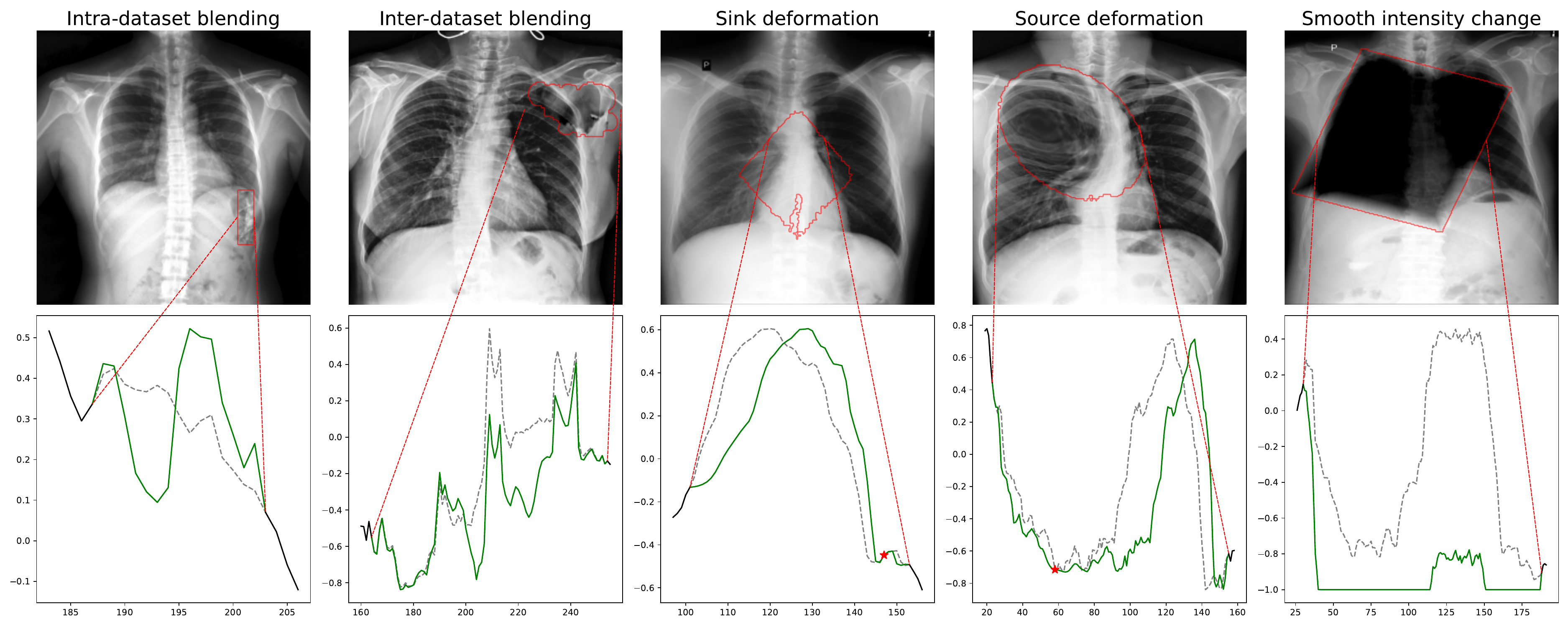}%
  }%
}
\setlength{\twosubht}{\ht\twosubbox}

\begin{figure}
    \centering
    \begin{minipage}{0.35\linewidth}
        \centering
         \includegraphics[height=\twosubht]{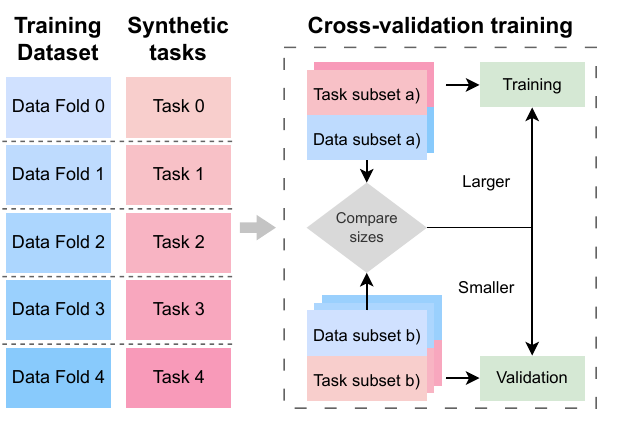}
    \caption{Our pipeline performs cross-validation over the synthetic task and data fold pairs.}
    \label{fig:cross_val}
    \end{minipage}\hfill
    \begin{minipage}{0.6\linewidth}
    \centering
    \includegraphics[height=\twosubht]{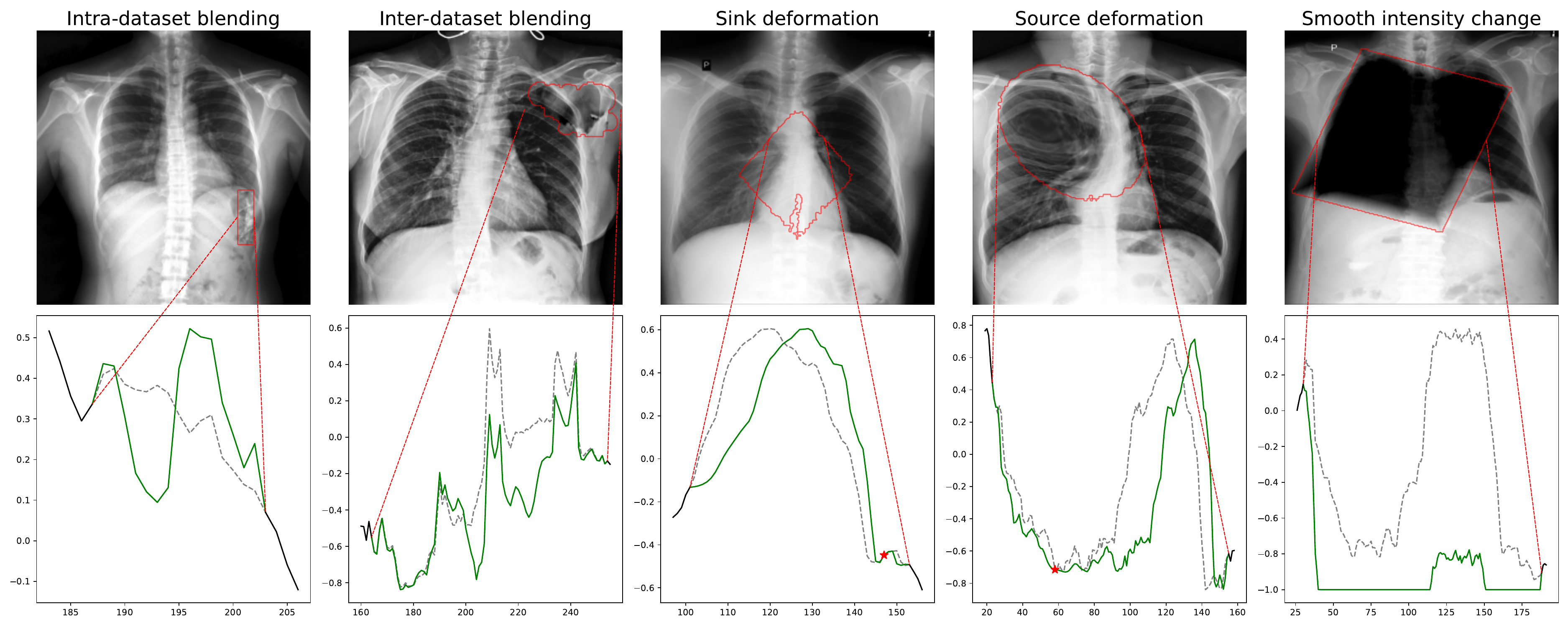}
    \caption{Examples of changes introduced by synthetic anomalies, showing the before (grey) and after (green) of a 1D slice across the affected area. \textcolor{red}{$\star$} - deformation centre for sink/source.}
    \label{fig:task_expl}
    \end{minipage}
\end{figure}

\noindent\textbf{Synthetic Tasks: }
Figure~\ref{fig:task_expl} shows examples of our self-supervised tasks viewed in both one and two dimensions.
Although each task produces visually distinct anomalies, they fall into three overall categories, based on blending, deformation, or intensity variation.
Also, all tasks share a common recipe: the target anomaly mask $M_h$ is always a randomly sized and rotated ellipse or rectangle (ellipsoids/cuboids in 3D); all anomalies are positioned such that at least 50\% of the mask intersects with the foreground of the image; and after one augmentation is applied, the process is randomly repeated (based on a fair coin toss, $p=0.5$), for up to a maximum of 4 anomalies per image.

\noindent\textbf{The intra-dataset blending task } 
Poisson image blending is the current state-of-the-art for synthetic anomaly tasks \cite{PII,NSA}, but it does not scale naturally to more than two dimensions or non-convex interpolation regions~\cite{morel2012fourier}.
Therefore, we extend \cite{perez2003poisson} and propose a D-dimensional variant of Poisson image editing following earlier ideas by~\cite{morel2012fourier}.

Poisson image editing~\cite{perez2003poisson} uses the image gradient to seamlessly blend a patch into an image.
It does this by combining the target gradient with Dirichlet boundary conditions to define a minimisation problem $\min_{f_{in}} \iint_{h} \left| \nabla f_{in} - \textbf{v} \right| ^2 $ with  $f_{in} |_{\partial h} =  f_{out} |_{\partial h}$, and $f_{in}$ representing the intensity values within the patch $h$. 
The goal is to find intensity values of $f_{in}$ that will match the surrounding values, $f_{out}$, of the destination image $x_i$, along the border of the patch and follow the image gradient, $\textbf{v} = \nabla \cdot = {\small \left( \frac{\partial \cdot}{\partial x} , \frac{\partial \cdot}{\partial y}  \right)}$, of the source image $x_j$.
Its solution is the Poisson equation $\Updelta f_{in} =  \textrm{div}\textbf{v} \textrm{ over } h$
with $ f_{in} \left|_{\partial h} =  f_{out} \right|_{\partial h}$.
Note that the divergence of $\textbf{v}$ is equal to the Laplacian of the source image $\Updelta x_j$.
Also, by defining $h$ as the axis-aligned bounding box of $M_h$, we can ensure the boundaries coincide with coordinate lines.
This enables us to use the Fourier transform method to solve this partial differential equation~\cite{morel2012fourier}, which yields a direct relationship between Fourier coefficients of $\Updelta f_{in}$ and $\textbf{v}$ after padding to a symmetric image.
To simplify for our use case, an image with shape $N_0\times\cdots\times N_{D-1}$, we replace the Fourier transformation with a discrete sine transform (DST) {\small$ \hat{f}_{\textbf{u}} = \sum_{d=0}^{D-1} \sum_{n=0}^{N_d-1} n \sin \left[\frac{\pi (n+1) (\textbf{u}_d+1) }{N_d+1} \right]$}.  
This follows as a DST is equivalent to a discrete Fourier transform of a real sequence that is odd around the zero-th and middle points, scaled by 0.5, which can be established for our images.
With this, the Poisson equation becomes congruent to a relationship of the coefficients, 
{\small
$\left( \sum_{d=0}^{D-1} \left(\frac{\pi (u_d+1) }{N_d+1}\right)^2 \right) \hat{f}_{\textbf{u}} \cong \sum_{d=0}^{D-1} \left(\frac{\pi (\textbf{u}_d+1) }{N_d+1}\right) \hat{\textbf{v}}_d $}
where $\textbf{v}{=}(\textbf{v}_0,..., \textbf{v}_{D-1})$ and $\hat{\textbf{v}}$ is the DST of each component. 
The solution for $\hat{f}_{\textbf{u}}$ can then be computed in DST space by dividing the right side through the terms on the left side and the destination image can be obtained through $x_i = DST^{-1}(\hat{f}_{\textbf{u}})$.
Because this approach uses a frequency transform-based solution, it may slightly alter areas outside of $M_h$ (where image gradients are explicitly edited) in order to ensure the changes are seamlessly integrated.
We refer to this blending process as $\Tilde{x} = PoissonBlend(x_i, x_j, M_h)$ in the following. The intra-dataset blending task therefore results from $\Tilde{x}_{intra} = PoissonBlend(x, x', M_h)$ with $x, x' \in \mathcal{D}$
with samples from a common dataset $\mathcal{D}$ and is therefore similar to the self-supervision task used in \cite{NSA} for 2D images.

\noindent\textbf{The inter-dataset blending task} follows the same process as intra-dataset blending but uses patches extracted from an external dataset $D'$, allowing for a greater variety of structures.
Therefore, samples from this task can be defined as $\Tilde{x}_{inter} = PoissonBlend(x, x', M_h)$ with $ x \in \mathcal{D}, x' \in \mathcal{D}'$.

\noindent\textbf{The sink/source tasks} shift all points in relation to a randomly selected deformation centre $c$.
For a given point $p$, we resample intensities from a new location $\Tilde{p}$.
To create a smooth displacement centred on $c$, we consider the distance $\lVert p - c\rVert_2$ in relation to the radius of the mask (along this direction), $d$.  
The extent of this displacement is controlled by the exponential factor $f>1$. 
For example, the sink task (Eqn.~\ref{eqn:sink}) with a factor of $f=2$ would take the intensity at $0.75d$ and place it at $0.5d$, effectively pulling these intensities closer to the centre.
Note that unlike the sink equation in \cite{melba:2022:013:tan} this formulation cannot sample outside of the boundaries of $P_{M_h}$ meaning it seamlessly blends into the surrounding area.
The source task (Eqn.~\ref{eqn:source}) performs the reverse, appearing to push the pixels away from the centre by sampling intensities towards it.
\begin{equation}
        \label{eqn:sink}
            \Tilde{x}_{p} = x_{\Tilde{p}}, \text{ }
        \Tilde{p} = c + d \frac{p-c}{\lVert p - c \rVert_2 } \left(1 - \left(1 - \frac{\lVert p - c \rVert_2}{d} \right)^f\right), c\in P_{M_h}, \text{ } \; \forall \; p \in P_{M_h}
\end{equation}

\begin{equation}
\Tilde{x}_{p} = x_{\Tilde{p}}, \text{ }
        \Tilde{p} = c + d \frac{p-c}{\lVert p - c \rVert_2 } \left(\frac{\lVert p - c \rVert_2}{d} \right)^f, \text{ }
  c\in P_{M_h}, \; \forall \; p \in P_{M_h}
  \label{eqn:source}
\end{equation}

\noindent\textbf{The smooth intensity change task} aims to either add or subtract an intensity over the entire anomaly mask.
To avoid sharp discontinuities at the boundaries, this intensity change is gradually dampened for pixels within a certain margin of the boundary.
This smoothing starts at a random distance from the boundary, $d_s$, and the change is modulated by $d_p/d_s$.

\noindent\textbf{Anomaly labelling: } 
In order to train and validate with multiple tasks simultaneously we use the same anomaly labelling function across all of our tasks.
The scaled logistic function, used in NSA \cite{NSA}, helps to translate raw intensity changes into more semantic labels.
But, it also rounds imperceptible differences up to a minimum score of about $0.1$.
This sudden and arbitrary jump creates noisy labels and can lead to unstable training.
We correct this semantic discontinuity by computing labels as $y = 1 - \frac{p_X(x)}{p_X(0)}$ with $X \sim \mathcal{N}(0, \sigma^2)$, instead of $y = \frac{1}{1 + e^{-k(x-x_0)}}$\cite{NSA}.
This flipped Gaussian shape is C1 continuous and smoothly approaches zero, providing consistent labels even for smaller changes.

\section{Experiments and Results}

\noindent\textbf{Data: }
We evaluate our method on T2-weighted brain MR and chest X-ray datasets to provide direct comparisons to state-of-the-art methods over a wide range of real anomalies.
For brain MRI we train on the Human Connectome Project (HCP) dataset \cite{VANESSEN20122222} which consists of 1113 MRI scans of healthy, young adults acquired as part of a scientific study.
To evaluate, we use the Brain Tumor Segmentation Challenge 2017
(BraTS) dataset \cite{bakas2018identifying}, containing 285 cases with either high or low grade glioma, and the ischemic stroke lesion segmentation challenge
2015 (ISLES) dataset \cite{MAIER2017250}, containing 28 cases with ischemic stroke lesions.
The data from both test sets was acquired as part of clinical routine.
The HCP dataset was resampled to have 1mm isotropic spacing to match the test datasets.
We apply z-score normalisation to each sample and then align the bounding box of each brain before padding it to a size of $160\times 224\times160$.
Lastly, samples are downsampled by a factor of two.

For chest X-rays we use the VinDr-CXR dataset \cite{nguyen2022vindr} including 22 different local labels.
To be able to compare with the benchmarks reported in \cite{cai2022dual-extension} we use the same healthy subset of 4000 images for training along with their test set (DDAD\textsubscript{ts}) of 1000 healthy and 1000 unhealthy samples, with some minor changes outlined as follows.
First note that \cite{cai2022dual-extension} derives VinDr-CXR labels using the majority vote of the 3 annotators.
Unfortunately, this means there are 52 training samples, where 1/3 of radiologists identified an anomaly, but the majority label is counted as healthy.
The same applies to 10 samples within the healthy testing subset.
To avoid this ambiguity, we replace these samples with leftover training data that all radiologists have labelled as healthy.
We also evaluate using the true test set (VinDr\textsubscript{ts}), where two senior radiologists have reviewed and consolidated all labels. 
For preprocessing, we clip pixel intensities according to the window centre and width attributes in each DICOM file, and apply histogram equalisation, before scaling intensities to the range $[-1,1]$.
Finally, images are resized to $256\times 256$.

\begin{figure}[t]
    \centering
    \includegraphics[width=\linewidth]{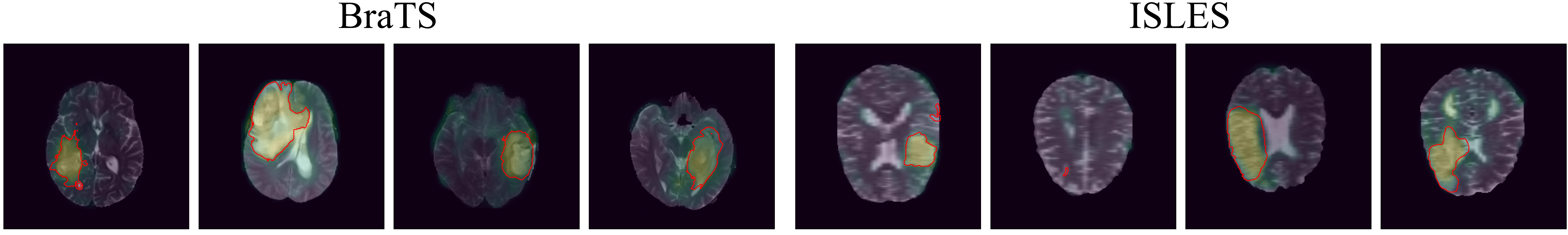}
    \caption{Examples of predictions on randomly selected BraTS and ISLES samples after training on HCP. The red contour outlines the ground truth segmentation.}
    \label{fig:brain_qual_examples}
\end{figure}

\noindent\textbf{Comparison to state-of-the-art methods: }
Validating on synthetic tasks is one of our main motivations; as such, we use a 1/4 (train/val.) task split to compare with benchmark methods.
For brain MRI, we evaluate results at the slice and voxel level, computing average precision (AP)
and area under the receiver operating characteristic curve (AUROC), as implemented in scikit learn~\cite{pedregosa2011scikit}.
Note that the distribution shift between training and test data (research vs. clinical scans) adds further difficulty to this task.
In spite of this, we substantially improve upon the current state-of-the-art (Table~\ref{tab:all_results} upper left).
In particular, we achieve a pixel-wise AP of 76.2 and 45.9 for BraTS and ISLES datasets respectively.
To make our comparison as faithful as possible, we also re-evaluate after post-processing our predictions to match the region and resolution used by CRADL, where we see similar improvement.
Qualitative examples are shown in Figure~\ref{fig:brain_qual_examples}.
Note that all baseline methods use a validation set consisting of real anomalous samples from BraTS and ISLES to select which anomaly scoring function to use.
We, however, only use synthetic validation data.
This further verifies that our method of using synthetic data to estimate generalisation works well.

\newcolumntype{R}{>{\raggedleft\arraybackslash}X}
\newcolumntype{Y}{>{\centering\arraybackslash}X}

\begin{table}[t]
    \centering
    \caption{\textbf{Upper left part:} Metrics on Brain MRI, evaluated on BraTS and ISLES, presented as AP/AUROC. $\stackrel{\text{\scalebox{.5}{CRADL setup}}}{\text{$\cdot$}}$ indicates that the metrics are evaluated over the same region and at the same resolution as CRADL~\cite{luth2023cradl}. 
    \textbf{Upper right part:} Metrics on VinDr-CXR, presented as AP/AUROC on the VinDr and DDAD test splits.
    $Random$ is the baseline performance of a random classifier. 
    \textbf{Lower part:} a sensitivity analysis of the average AP of each individual fold (mean$\pm$s.d.) alongside that of the model ensemble, varying how many tasks we use for training versus validation. Best results are highlighted in bold.}
    \resizebox{\linewidth}{!}{
    \begin{tabular}{lccccp{0.01cm}ccp{0.01cm}ccp{0.01cm}cc}
    \cmidrule(r){1-9}\cmidrule(r){10-14}
    &&& \multicolumn{5}{c}{Brain MRI} & & \multicolumn{5}{c}{Chest X-Ray (CXR)} \\
    \cmidrule(r){4-8}\cmidrule(r){10-14}
    &&& \multicolumn{2}{c}{Slice-wise} & & \multicolumn{2}{c}{Pixel-wise} &  & \multicolumn{2}{c}{Sample-wise} & & \multicolumn{2}{c}{Pixel-wise} \\
    &&& BraTS17 & ISLES &  & BraTS & ISLES & & DDAD\textsubscript{ts} & VinDr\textsubscript{ts} && DDAD\textsubscript{ts} & VinDr\textsubscript{ts}\\
    \cmidrule(r){4-5}\cmidrule(r){7-8}\cmidrule(r){10-11}\cmidrule(r){13-14}
    \\[-0.4cm]
     & & &  &  & &  &  & & \multicolumn{1}{|c}{59.8/55.8} & \multicolumn{3}{r}{MemAE \cite{gong2019memorizing}} &  \multicolumn{1}{l}{\multirow{5}{*}{\rotatebox{-90}{Methods CXR}}}\\
    \multirow{5}{*}{\rotatebox{90}{Methods MRI}} &\multicolumn{2}{l}{VAE}   & 80.7/83.3 & 51.9/71.7 & & 29.8/92.5 & 7.7/87.5 & &\multicolumn{1}{|c}{74.8/76.3}& \multicolumn{3}{r}{f-AnoGAN \cite{SCHLEGL201930}}\\
     &\multicolumn{2}{l}{ceVAE\cite{zimmerer2018context}} & 85.6/86.5 & 54.1/72.7 && 48.3/94.8 & 14.5/87.9 & &\multicolumn{1}{|c}{72.8/73.8} &\multicolumn{3}{r}{AE-U \cite{10.1007/978-3-030-59725-2_51}}\\
     &\multicolumn{2}{l}{CRADL\cite{luth2023cradl}} & 81.9/82.6 & 54.9/69.3 & & 38.0/94.2 & 18.6/89.8 & &\multicolumn{1}{|c}{49.9/48.2}&\multicolumn{3}{r}{FPI \cite{melba:2022:013:tan}}\\    
     &\multicolumn{2}{l}{$\stackrel{\text{\scalebox{.5}{CRADL setup}}}{\text{Ours}}$} & 87.6/89.4 &  61.3/80.2 && 76.2/98.7 &  46.5/97.1 & &\multicolumn{1}{|c}{65.8/65.9}&\multicolumn{3}{r}{PII \cite{PII}}\\
     &\multicolumn{2}{l}{$\stackrel{\text{\scalebox{.5}{CRADL setup}}}{Random}$} & 49.0/50.0 & 36.6/50.0 && 2.4/50.0 & 1.1/50.0 & &\multicolumn{1}{|c}{65.8/64.4}&\multicolumn{3}{r}{NSA \cite{NSA}}\\
     \multicolumn{3}{l}{$Random$} &  40.3/50.0 &  29.4/50.0 &&  1.7/50.0 &  0.8/50.0 && 50.0/50.0 & 31.6/50.0 && 4.5/50.0 &  2.7/50.0\\
     \multicolumn{3}{l}{Ours} & \textbf{87.6/92.2} &  \textbf{62.0/84.6} & &  \textbf{76.2/99.1} &  \textbf{45.9/97.9} && \textbf{78.4/76.6} &  \textbf{71.2/81.1} &&  \textbf{21.1/75.6} &  \textbf{21.4/81.2}\\
  \cmidrule(r){1-9}\cmidrule(r){10-14}
  \\[-0.4cm]
  \cmidrule(r){1-9}\cmidrule(r){10-14}

    \multirow{8}{*}{\rotatebox{90}{Train/val. split $abl.$}}& \multirow{2}{*}{1/4} & all &  83.4$\pm$4.4 &  59.3$\pm$2.2 &&  46.9$\pm$14.9 &  23.7$\pm$7.7 &&  74.7$\pm$4.9 &  66.3$\pm$4.4 &&  15.3$\pm$3.7 &  15.2$\pm$4.5 \\
    
    &  & ens.  &\textbf{87.6} &          \textbf{62.0} &&           \textbf{76.2} &          \textbf{45.9} &&          78.4 &          71.2 &&          21.1 &          21.4 \\
    
    & \multirow{2}{*}{2/3} & all & 82.5$\pm$3.3 &  55.9$\pm$8.5 &&  42.8$\pm$12.8 &  21.2$\pm$9.3 &&  78.6$\pm$1.4 &  71.0$\pm$1.4 &&  19.2$\pm$1.7 &  19.5$\pm$1.8 \\
    
    &  & ens. & 85.7 &          58.4 && 72.2 & 41.0 && \textbf{80.7} & \textbf{73.8} && 24.0 &          \textbf{24.7} \\
    
    & \multirow{2}{*}{3/2} & all &  81.1$\pm$4.3 &  52.5$\pm$4.7 && 37.9$\pm$11.1 &  15.4$\pm$3.3 & & 78.7$\pm$1.8 &  71.1$\pm$1.4 && 20.3$\pm$1.4 &  20.4$\pm$1.7 \\
    
    &  & ens.  &  84.0 & 55.0 && 63.7 & 26.6 && 80.4 & 73.3 && \textbf{24.3} & \textbf{24.7}\\
    
    & \multirow{2}{*}{4/1} & all & 81.5$\pm$2.7 & 53.1$\pm$2.3 && 36.1$\pm$9.0 & 16.5$\pm$5.0 && 79.2$\pm$1.3 & 71.8$\pm$1.3 && 20.4$\pm$0.9 & 21.1$\pm$0.9 \\
    & & ens. & 83.1 & 54.7 && 52.5 & 23.7 && 80.5 & 73.6 && 23.5 & 24.5 \\
    
    \bottomrule
    \end{tabular}
    }
    \label{tab:all_results}
\end{table}

For both VinDr-CXR test sets we evaluate at a sample and pixel level, although previous publications have only reported their results at a sample level.
We again show performance above the current state-of-the-art (Table~\ref{tab:all_results} upper right).
Our results are also substantially higher than previously proposed self-supervised methods, improving on the current state-of-the-art NSA~\cite{NSA} by 12.6 to achieve 78.4 image-level AP.
This shows that our use of synthetic validation data succeeds where their fixed training schedule fails.

\noindent\textbf{Ablation and sensitivity analysis on cross-validation structure: }
We also investigate how performance changes as we vary the number of tasks used for training and validation (Table~\ref{tab:all_results} lower).
For VinDr-CXR, in an individual fold, the average performance increases as training becomes more diverse (i.e. more tasks); however, the performance of the ensemble plateaus.
Having more training tasks can help the model to be sensitive to a wider range of anomalous features.
But as the number of training tasks increases, so does the overlap between different models in the ensemble, diminishing the benefit of pooling predictions. 
This could also explain why the standard deviation (across folds) decreases as the number of training tasks increases, since the models are becoming more similar.
Our best configuration is close to being competitive with the state-of-the-art \textit{semi}-supervised method DDAD-ASR~\cite{cai2022dual-extension}.
Even though their method uses twice as much training data, as well as some real anomalous data, our purely synthetic method begins to close the gap (AP of \cite{cai2022dual-extension} 84.3 vs. ours 80.7 on DDAD\textsubscript{ts}).
For the brain datasets, all metrics generally decrease as the number of training tasks increases.
This could be due to the distribution shift between training and test data.
Although more training tasks may increase sensitivity to diverse irregularities, this can actually become a liability if there are differences between (healthy) training and test data (e.g. acquisition parameters).
More sensitive models may then lead to more ``false'' positives.

\noindent\textbf{Discussion: }
We demonstrate the effectiveness of our method in multiple settings and across different modalities.
A unique aspect of the brain data is the domain shift.
The HCP training data was acquired at a much higher isotropic resolution than the BraTS and ISLES test data, which are both anisotropic. 
Here we achieve the best performance using more tasks for validation, which successfully reduces overfitting and hypersensitivity.
Incorporating greater data augmentations, such as simulating anisotropic spacing, could further improve results by training the model to ignore these transformations.
We also achieve strong results for the X-ray data, although precise localisation remains a challenging task.
The gap between current performance and clinicially useful localisation should therefore be high priority for future research.

\section{Conclusion}
In this work we use multiple synthetic tasks to both train and validate self-supervised anomaly detection models.
This enables more robust training without the need for real anomalous training \textit{or} validation data.
To achieve this we propose multiple diverse tasks, exposing models to a wide range of anomalous features.
These include patch blending, image deformations and intensity modulations.
As part of this, we extend Poisson image editing to images of arbitrary dimensions, enabling the current state-of-the-art tasks to be applied beyond just 2D images.
In order to use all of these tasks in a common framework we also design a unified labelling function, with improved continuity for small intensity changes.
We evaluate our method on both brain MRI and chest X-rays and achieve state-of-the-art performance and above.
We also report pixel-wise results, even for the challenging case of chest X-rays.
We hope this encourages others to do the same, as accurate localisation is essential for anomaly detection to have a future in clinical workflows. 

\noindent\textbf{Acknowledgements:}
We thank EPSRC for DTP funding and HPC resources provided by the Erlangen National High Performance Computing Center (NHR @ FAU) of the Friedrich-Alexander-Universität Erlangen-Nürnberg (FAU) under the NHR project b143dc. NHR funding is provided by federal and Bavarian state authorities. NHR@FAU hardware is partially funded by the German Research Foundation (DFG) – 440719683. Support was also received by the ERC - project MIA-NORMAL 101083647 and DFG KA 5801/2-1, INST 90/1351-1.

%
%
%
\bibliographystyle{splncs04}
\bibliography{mybibliography}

\newpage








\begin{figure}[t]
    \centering
    \includegraphics[width=\linewidth]{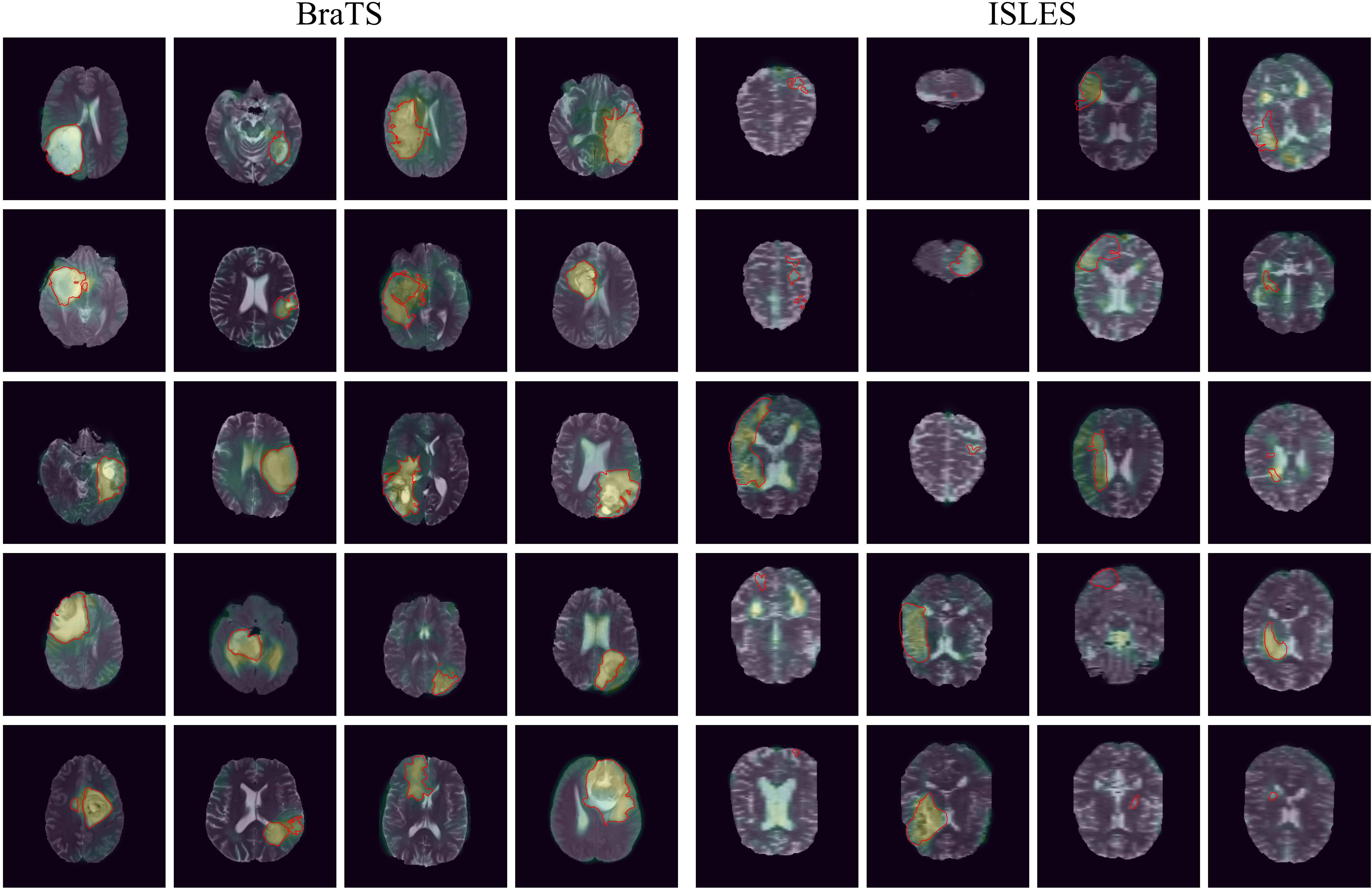}
    \caption{More randomly selected predictions on BraTS and ISLES samples after training on HCP. The red contour outlines the ground truth segmentation.}
    \label{fig:brain_appendix_examples}
\end{figure}

\begin{figure}[t]
    \centering
    \includegraphics[width=0.9\linewidth]{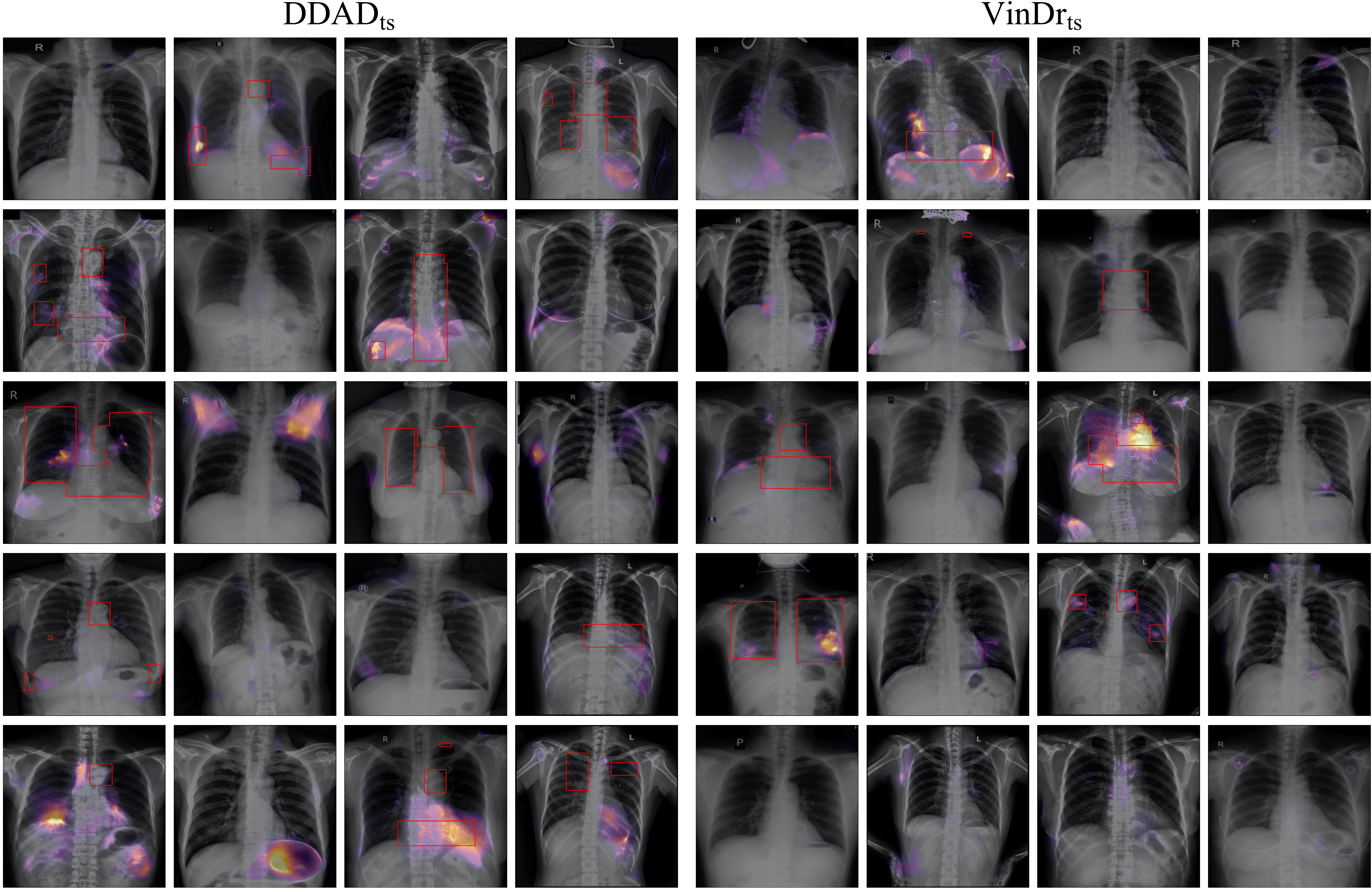}
    \caption{Randomly selected predictions on DDAD\textsubscript{ts} and VinDr\textsubscript{ts} samples after training on VinDr. The red contour outlines the ground truth segmentation. Predictions scaled by a factor of 2 for visualisation purposes. The model generally underpredicts, with some exceptions such as the raised arms in the third row, second column, which could be viewed as a logical anomaly. }
    \label{fig:vindr_appendix_examples}
\end{figure}

\begin{table}
    \centering
    \caption{Metrics of all models evaluated on their respective test sets, presented as AP/AUROC. ens. - ensemble.}
    \resizebox{0.95\linewidth}{!}{
    \begin{tabular}{llccccccccccc}
\toprule
   Train &          &  \multicolumn{5}{c}{Brain MRI} & & \multicolumn{5}{c}{Chest X-Ray (CXR)} \\
    \cmidrule(lr){3-7}\cmidrule(r){9-13}
   val. & Model  & \multicolumn{2}{c}{Slice-wise} & & \multicolumn{2}{c}{Pixel-wise} & & \multicolumn{2}{c}{Sample-wise} & & \multicolumn{2}{c}{Pixel-wise} \\
    split &      &      BraTS &      ISLES & &      BraTS &      ISLES &  & DDAD\textsubscript{ts} & VinDr\textsubscript{ts} & &  DDAD\textsubscript{ts} & VinDr\textsubscript{ts} \\
\midrule
\multirow{6}{*}{1/4} & 0 &  80.5/88.6 &  56.5/80.9 & &  32.2/96.5 &  12.0/94.7 & &   79.1/77.3 &  72.0/81.5 & &  19.7/73.1 &  20.9/79.8 \\
    & 1 &  83.1/89.7 &  59.2/82.7 & &  59.6/98.3 &  29.2/97.2 & &   78.7/76.2 &  69.5/79.4 & &  17.4/69.8 &  18.6/73.8 \\
    & 2 &  86.2/91.7 &  58.2/82.0 & &  29.1/92.5 &  20.0/93.9 & &   67.3/66.5 &  61.3/73.4 & &  10.9/61.4 &  10.2/65.5 \\
    & 3 &  89.1/92.6 &  62.4/84.8 & &  57.8/98.0 &  30.3/96.9 & &   72.3/71.6 &  63.3/76.3 & &  12.3/64.6 &  11.8/69.9 \\
    & 4 &  78.2/87.0 &  60.4/83.0 & &  55.6/97.2 &  27.1/93.7 & &   75.8/72.9 &  65.2/76.8 & &  16.2/64.5 &  14.4/66.9 \\
    \cmidrule{2-13}
    & ens. &  87.6/92.2 &  62.0/84.6 & &  76.2/99.1 &  45.9/97.9 & &   78.4/76.6 &  71.2/81.1 & &  21.1/75.6 &  21.4/81.2 \\
\midrule
\multirow{11}{*}{2/3} & 0 &  83.1/89.8 &  58.9/83.2 & &  53.5/97.9 &  20.8/96.2 & &   81.0/79.4 &  72.9/82.8 & &  19.8/72.9 &  21.3/78.9 \\
    & 1 &  79.6/86.6 &  40.9/70.4 & &  21.8/94.7 &   8.1/92.2 & &   77.4/76.3 &  71.0/81.6 & &  21.0/76.4 &  20.5/80.7 \\
    & 2 &  84.0/89.6 &  55.0/80.0 & &  29.2/95.5 &  14.9/95.5 & &   78.7/76.7 &  72.8/82.4 & &  21.5/77.4 &  21.4/83.8 \\
    & 3 &  80.6/88.2 &  57.5/82.0 & &  48.7/97.9 &  20.4/95.7 & &   78.9/77.8 &  68.6/79.9 & &  19.3/71.6 &  19.3/78.1 \\
    & 4 &  84.4/89.9 &  51.5/77.4 & &  39.1/96.3 &  14.4/94.3 & &   79.5/78.4 &  71.7/82.2 & &  19.0/70.1 &  21.1/77.5 \\
    & 5 &  82.4/88.5 &  46.1/74.4 & &  30.0/96.1 &  13.2/95.2 & &   79.2/77.5 &  71.0/81.2 & &  19.4/69.4 &  20.1/76.0 \\
    & 6 &  80.9/88.5 &  60.1/82.9 & &  55.4/97.8 &  24.4/95.2 & &   79.0/76.8 &  69.9/81.3 & &  17.9/69.8 &  18.0/76.6 \\
    & 7 &  90.2/93.4 &  71.8/86.6 & &  58.2/97.8 &  40.1/95.6 & &   75.9/75.1 &  69.9/80.4 & &  15.9/72.0 &  15.8/78.1 \\
    & 8 &  81.1/88.3 &  60.7/80.6 & &  53.6/97.2 &  29.1/94.9 & &   77.5/75.8 &  70.0/80.7 & &  18.1/71.2 &  17.7/77.8 \\
    & 9 &  78.9/87.5 &  56.3/80.0 & &  38.6/97.1 &  26.9/96.1 & &   79.2/77.6 &  71.5/81.6 & &  20.4/73.3 &  19.6/80.4 \\
    \cmidrule{2-13}
    & ens. &  85.7/90.7 &  58.4/81.7 & &  72.2/98.7 &  41.0/97.2 & &   80.7/79.5 &  73.8/83.6 & &  24.0/79.9 &  24.7/86.0 \\
\midrule
\multirow{11}{*}{3/2} & 0 &  80.2/87.2 &  46.9/73.3 & &  37.9/96.5 &  10.8/94.4 & &   78.4/77.3 &  71.7/81.2 & &  21.5/75.4 &  21.7/80.6 \\
    & 1 &  82.1/89.5 &  57.5/82.1 & &  29.9/96.5 &  16.6/96.1 & &   80.1/78.7 &  72.0/82.1 & &  20.2/73.2 &  21.1/79.8 \\
    & 2 &  81.6/88.8 &  58.6/82.5 & &  44.6/98.0 &  19.7/96.3 & &   80.6/79.8 &  72.0/82.2 & &  20.5/74.2 &  20.6/81.1 \\
    & 3 &  88.4/92.3 &  49.5/76.7 & &  54.0/95.5 &  18.6/94.3 & &   79.6/78.2 &  73.0/82.9 & &  22.1/76.7 &  22.9/83.6 \\
    & 4 &  77.9/85.4 &  45.7/73.2 & &  51.6/97.4 &  19.0/94.9 & &   80.3/78.9 &  71.6/81.9 & &  21.3/73.1 &  21.3/80.0 \\
    & 5 &  81.3/88.1 &  55.5/80.7 & &  27.9/96.8 &  15.0/95.8 & &   78.2/76.5 &  70.0/80.2 & &  21.2/76.8 &  20.6/82.8 \\
    & 6 &  87.3/91.6 &  57.1/80.0 & &  49.6/97.2 &  17.6/94.4 & &   78.3/76.9 &  71.3/81.9 & &  19.4/71.4 &  19.9/76.1 \\
    & 7 &  79.0/86.8 &  48.2/75.4 & &  31.8/96.7 &  12.1/94.0 & &   76.0/75.0 &  68.5/79.4 & &  17.9/70.8 &  17.6/76.3 \\
    & 8 &  79.6/87.3 &  53.8/78.5 & &  25.3/96.5 &  11.7/95.1 & &   80.1/78.6 &  71.2/82.3 & &  20.6/74.2 &  20.5/80.8 \\
    & 9 &  73.9/84.4 &  52.4/76.8 & &  26.6/95.1 &  13.4/94.2 & &   75.4/73.6 &  69.2/79.1 & &  18.3/72.1 &  17.7/79.2 \\
    \cmidrule{2-13}
    & ens. &  84.0/89.6 &  55.0/79.7 & &  63.7/98.5 &  26.6/96.8 & &   80.4/79.4 &  73.3/83.1 & &  24.3/80.1 &  24.7/85.9 \\
\midrule
\multirow{6}{*}{4/1} & 0 &  84.2/89.8 &  50.2/76.1 & &  37.4/97.3 &  13.1/95.4 & &   78.8/76.9 &  72.2/81.9 & &  20.5/72.1 &  21.8/78.5 \\
    & 1 &  79.8/87.4 &  53.3/79.4 & &  35.2/96.7 &  20.4/96.0 & &   79.9/78.2 &  72.2/82.5 & &  19.9/75.0 &  22.0/83.0 \\
    & 2 &  79.8/87.6 &  55.2/80.6 & &  24.8/96.3 &  11.4/95.1 & &   81.0/79.3 &  73.5/82.9 & &  21.8/77.1 &  21.2/81.7 \\
    & 3 &  84.5/89.7 &  51.3/76.3 & &  49.6/97.9 &  23.0/95.7 & &   78.6/77.6 &  70.4/80.6 & &  20.5/74.8 &  20.7/82.5 \\
    & 4 &  79.0/87.4 &  55.4/78.8 & &  33.3/97.2 &  14.6/95.5 & &   77.7/76.2 &  70.5/80.8 & &  19.5/71.7 &  19.9/78.6 \\
    \cmidrule{2-13}
    & ens. &  83.1/89.2 &  54.7/79.6 & &  52.5/98.4 &  23.7/96.8 & &   80.5/79.3 &  73.6/83.3 & &  23.5/78.8 &  24.5/85.5 \\
    \bottomrule
    \end{tabular}
    }

    \label{tab:all_individual_results}
\end{table}


\begin{table}[]
\caption{Experiment hyperparameters. Labeller $\sigma$ is used to parameterise the anomaly labelling function, external blending dataset is used for inter-dataset blending task.}
    \resizebox{\linewidth}{!}{
\begin{tabular}{lccccccc}
\toprule
\multirow{2}{*}{Dataset}   & Labeller  & Learning &U-Net  & External Blending & Batch & GPU & Average training\\
& $\sigma$ & rate & Depth & Dataset & size & footprint & time  (hours)\\
\midrule
Brain MRI & 0.2                  & 1e-3               & 5           & MOOD Abdomen              & 8          & 20 GB                     & 15.17                     \\
VinDr-CXR & 0.2                  & 5e-3               & 7           & Imagenet                  & 50         & 18 GB                     & 10.09                \\
\bottomrule
\end{tabular}
}
\end{table}


\end{document}